\documentclass{article}
\usepackage[preprint]{neurips_2019}

\usepackage[utf8]{inputenc} 
\usepackage[T1]{fontenc}    
\usepackage{hyperref}       
\usepackage{url}            
\usepackage{booktabs}       
\usepackage{amsfonts}       
\usepackage{nicefrac}       
\usepackage{microtype}      
\usepackage{verbatim}
\usepackage{xcolor}
\usepackage{subcaption}
\usepackage{caption}
\usepackage{enumitem}
\usepackage{amsmath}
\usepackage{shortbold}

\usepackage{nicefrac} 

\usepackage{placeins}

\DeclareMathAlphabet\mathbfcal{OMS}{cmsy}{b}{n}

\newif\ifincludecomment
\includecommenttrue 

\ifincludecomment
  
\else
  \NewEnviron{guidance}{}{}
\fi

\usepackage[colorinlistoftodos,textsize=tiny, textwidth=18mm]{todonotes}
\ifincludecomment
\newcommand{\maybecomment}[1]{\todo[color=olive!40]{#1}} 
 
\newcommand{\maybedelete}[1]{\todo[color=blue!40]{#1}}

\else
  \newcommand{\maybecomment}[1]{}
  
\newcommand{\maybedelete}[1]{} 
\fi

\title{Separable Layers Enable Structured Efficient Linear Substitutions}

\author{%
  Gavin Gray, Elliot J. Crowley, Amos Storkey\\
  School of Informatics\\
  University of Edinburgh\\
    \texttt{\scriptsize{\{g.d.b.gray, elliot.j.crowley, a.storkey\}@ed.ac.uk}} \\
}

\begin{document}

\maketitle

\begin{abstract}
In response to the development of recent efficient dense layers, this
    paper shows that something as simple as replacing linear components in
    pointwise convolutions with structured linear decompositions also produces
    substantial gains in the efficiency/accuracy tradeoff. Pointwise
    convolutions are fully connected layers and are thus prepared for
    replacement by structured transforms.  Networks using such layers are
    able to learn the same tasks as those using standard convolutions, and
    provide Pareto-optimal benefits in efficiency/accuracy, both in terms of
    computation (mult-adds) and parameter count (and hence memory). Code is available at \url{https://github.com/BayesWatch/deficient-efficient}.
\end{abstract}

\section{Introduction}
\label{intro}

There is continued and pressing need for efficient implementations of neural
networks, for example, in real-time processing, and for deployment on embedded
devices. Efficiency may be achieved, for example, by inducing sparsity in the trained
network~\citep{lecun1989optimal} or quantisation of the weights being
used~\citep{courbariaux2015binaryconnect}. Significant redundancy in the
trained parameters used in neural networks allows
this~\citep{alvarez2016learning}.

A number of methods have reduced stored size or computational cost in neural networks by providing efficient alternatives to fully connected layers; these include
HashedNet~\citep{chen2015compressing},
Tensor-Train~\citep{novikov2015tensorizing} and ACDC~\citep{moczulski2016acdc}.
However, as modern neural networks largely avoid
fully connected layers, these are not typically considered in a convolutional network setting. Yet, the
pointwise convolution in current depthwise-separable layers is a full
matrix multiply. In this paper we ask whether methods developed for making fully connected layers efficient provide an effective approach for improving efficiency of convolutional neural networks.

A pointwise convolution is a $1 \times 1$ convolution applied at every spatial
point on the input tensor; equivalent to a dense layer treating each point as
an independent example.  Pointwise convolutions consume the vast majority of
the parameter and computation budget in competitive image classification
architectures~\citep{huang2017densely,zoph2017neural,liu2019darts}. In this
paper we assess the benefit of substituting a number of structured efficient
alternatives in place of these layers and compare the efficiency/accuracy
tradeoff on standard benchmark image classification problems.

The main contributions of this paper are:
\begin{itemize}
    \item A comparison of a number of approaches for replacing dense
        layers that occur within convolutional blocks (e.g.\ pointwise layers in separable convolutions), in terms of their efficiency.
    \item A principled method to tune weight decay when using compressed layers that maintains stability, making substitution more practical.
    \item Demonstration that the use of these methods provides Pareto-optimal networks, as demonstrated on CIFAR-10, along with characterisation of the compression rates on ImageNet.
    \item An indication of the best performing and most stable alternative layer parameterisations, to direct practitioners and researchers towards the best network choices.
\end{itemize}

\section{Related Work}
\label{relatedwork}

Approaches for structured efficient linear layers (SELLs), such as
ACDC~\citep{moczulski2016acdc}, differ from other work in
efficiency in that they provide a \emph{substitute} to a fully connected
layer, but are trained using the same methods that would otherwise be used (e.g.\ stochastic gradient descent). Other approaches (e.g.~\citealp{alvarez2017compression}) provide an
alternative training algorithm for such layers.

In this paper we compare candidate substitute layers; empirically we provide evidence of state-of-the-art performance of some of these approaches on modern fully convolutional networks, which are substantially different from the setting in which many of these methods were proposed. For example, work on SELLs focused on the replacement of fully connected layers in
AlexNet~\citep{moczulski2016acdc,yang2015deep}, which now has little relevance
to modern convolutional architectures.  The ACDC
layer~\citep{moczulski2016acdc} was inspired by circulant transforms and
subsequent work has focused on the potential of these
transforms~\citep{zhao2019building}.  Circulant transforms and SELLs were later
unified under low displacement rank (LDR)
transformations~\citep{zhao2017ldr,sa2018two}.  Also, circulant transforms have
been defined as a special case of unitary group convolution (UGConv) layers~\citep{zhao2019building},
allowing a relation to be drawn to the successful ShuffleNet
architecture~\citep{zhang2018shuffle}.

To summarise, we present strong empirical results on contemporary
image classification problems for members of a number of classes of transformations:
\begin{itemize}
    \item Low-rank substitutions~\citep{alvarez2016learning,alvarez2017compression,sainath2013low,jaderberg2014speeding,novikov2015tensorizing,kossaifi2017tensor,kim2015compression}.
    \item SELLs~\citep{yang2015deep,moczulski2016acdc,bojarski2016structured}.
    \item Circulant and Block-Circulant layers~\citep{liao2019circ,zhao2019building,cheng2015exploration,araujo2019the}.
    \item Toeplitz-like transforms~\citep{lu2016learning,sindhwani2015structured}.
    \item ShuffleNet-like layers~\citep{zhang2018shuffle,gao2018channelnets}.
    \item Generalised LDR methods~\citep{thomas2018learning,sa2018two}.
    \item Weight hashing schemes~\citep{chen2015compressing}.
\end{itemize}

\section{Methods}
\label{methods}

All of the compressed convolutions in this paper are linear reparameterisations of standard convolutions; we demonstrate that when we replace the convolutional layers of modern networks with these compressed reparameterisations, we can achieve high performance for reduced parameter counts and computational cost.

The various substitutions considered are described in
Section~\ref{methods:linear}. Each either uses fewer parameters, fewer
mult-adds or both. These are by no means exhaustive, and have been selected to cover many of the different transformation classes listed in Section~\ref{relatedwork}.

To make training these layers practical, we have to account for the effect of
substituting a compressed convolution on the choice for weight decay.  We derive the weight decay parameters needed to stabilise training in
Section~\ref{methods:crs}.  

\subsection{Substitute Efficient Linear Transforms}
\label{methods:linear}

Here, we describe each of the methods being
compared. All provide an approximation to the operation of a dense
random matrix in a linear layer: a matrix-vector product of that matrix
with an input vector of the form $\yB = \WB \xB$, where $\yB$ is the output vector, $\WB$ is the dense random matrix, and $\xB$ is the input vector.

\subsubsection{ACDC}
In~\cite{moczulski2016acdc}, $\WB$ is decomposed into a stack of $L$ ACDC layers:

\begin{align}
	\WB = \prod_{l=1}^L \AB_l \CB \DB_l \CB^{-1} \PB
    \label{eq:acdc}
\end{align}
where $\AB$ and $\DB$ are diagonal matrices, $\CB$ and $\CB^{-1}$ are forward and inverse discrete cosine transforms, and $\PB$ is a random permutation matrix. As the operation of a random permutation may be time consuming, we replace $\PB$ with a \emph{riffle shuffle}. A riffle shuffle is a fixed permutation, splitting the input in half and
then interleaving the two halves~\citep{gilbert1955theory}.  This was found to work as well as a
fixed random permutation and can be evaluated much faster as observed
by~\citet{zhang2018shuffle}. For $\WB \in \mathbb{R}^{N \times N}$ the computational complexity is $O (N \log N)$
and storage cost is $O(N)$~\citep{moczulski2016acdc}.

\subsubsection{Tensor-Train}
\newcommand{\AT}{\mathbfcal{A}}

First, we assume it is possible to map a higher dimensional tensor to our weight matrix using a reshape operation: $ \yB = \WB \xB = \text{reshape}^{\mathbb{R} \in N \times N}(\AT) \xB.$ This allows us to use a tensor decomposition to represent $\AT$ and implement
the linear transform using fewer parameters. In Tensor-Train (TT)~\citep{oseledets2011tensor},  $\AT$ is decomposed as:

\begin{align}
    \AT(i_i, ..., i_d) = \GB_1(i_1) ... \GB(i_d)
\end{align}

Where $\GB_k (i_k)$ are $r_{k-1} \times r_k$ matrices, with the boundary 
conditions that ensure $r_0 = r_d = 1$. Each element of the tensor can then be
reproduced by performing this sequence of matrix products.

The parameter savings using this method depend on the number of dimensions
possessed by the tensor storing the weights. It
is possible to perform a matrix-vector, or matrix-matrix, product between
two TT tensors. Alternatively, as in our experiments, we can compute the weight matrix from the $\GB_k$ factors and backpropagate the error to update those factors with automatic differentiation.

In our experiments we found it
best to reshape weight matrices to 3 dimensions, with approximately equal
sizes. We then set the TT-rank $r_1, ..., r_{d-1}$ to control the level
of compression. This is in line with previous work substituting TT tensors
into fully connected layers of deep neural networks~\citep{novikov2015tensorizing}.

\subsubsection{Tucker decomposition}

\newcommand{\GT}{\mathbfcal{G}}

The Tucker decomposition also decomposes a tensor
$\AT \in \mathbb{R}^{I_0, ..., I_d}$, but in this case uses a low rank core $\GT
\in \mathbb{R}^{R_0, ..., R_d}$ projected by factors $\UB_k \in \mathbb{R}^{R_k,
I_k}$~\citep{kim2015compression}:

\begin{align}
    \AT = \GT \times_0 \UB_0 ... \times_d \UB_d    
\end{align}

Where $\times_k$ denotes the \emph{$k$-mode product}:
a matrix product on dimension $k$ while casting over the remaining dimensions.
In our experiments, to compare with TT, we only use the 
Tucker decomposition to store our weight matrices. As with
the TT decomposition, we compute the weight matrix, then
backpropagate gradients in order to update the $\UB_k$ 
factors.

\subsubsection{Rank-factorised (RF) decomposition}

\newcommand{\din}{d_{\text{in}}}
\newcommand{\dbn}{d_{\text{bn}}}
\newcommand{\dout}{d_{\text{out}}}

A rank factorised matrix is a linear bottleneck. It is chosen as a
baseline against which to compare methods from the literature.
In place of a dense matrix we first map an input to a smaller number of
dimensions, and then back to the output number of dimensions. This uses two
weight matrices $\WB_1 \in \mathbb{R}^{\dbn \times \din}$ and $\WB_2 \in
\mathbb{R}^{\dout \times \dbn}$, where the input dimensionality is $\din$,
bottleneck is $\dbn$ and output is $\dout$. The linear transformation from an
input $\XB$ to an output $\YB$ can then be expressed as $\yB = \WB \xB = \WB_2 \WB_1 \xB$. 

This parameterisation can be implemented in popular deep learning frameworks
with two linear layers in sequence, and can give significant efficiency benefits. For $\din=\dout=d$, the number of parameters used by applying
a dense weight matrix $\WB$ to an input vector is $d^2$, while
the total parameters used in $\WB_1$ and $\WB_2$ is $\frac{2 d^2}{b}$ where $b = \frac{d}{\dbn}$.

\subsubsection{HashedNet}

A virtual weight matrix $\VB$ is built from real weights $\wB$ using a hash
function $\hB$ to index those weights:

\begin{align}
    y_i = \sum_{j=1}^N V_{ij} x_j = \sum_{j=1}^N w_{h(i,j)} x_j
    \label{eq:hashed}
\end{align}

HashedNets use a hash function to retrieve the weights used in their
network~\citep{chen2015compressing}. The particular hash function used in this
case takes as input indices in the \emph{virtual} weight matrix, $\VB$, used in
the linear transformation, and produces as output a single index into a set of
\emph{real} weights $\wB$. As the compression only depends on the number of
virtual weights we choose to use, this method is extremely flexible for storage compression. Note that some virtual weights may not be used, as described in Appendix~\ref{appendix:hashednet}.

\subsubsection{Linearised ShuffleNet}

The Shufflenet unit~\citep{zhang2018shuffle} can be related to circulant
transforms like ACDC~\citep{moczulski2016acdc} by defining a generalised UGConv
block~\citep{zhao2019building}. Unlike applications of circulant transforms,
these units are used to implement a state-of-the-art image classification
architecture (ShuffleNet). We propose a linear version of the unit, represented by:

\begin{align}
    \yB = \WB \xB = \BB_2 \PB \BB_1 \xB
\end{align}
\newcommand{\DT}{\mathbfcal{D}}
where $\BB_1$ and $\BB_2$ are block diagonal matrices implemented by grouped
$1 \times 1$ convolutions, and $\PB$ is a permutation implemented by a 
riffle shuffle. While this was not proposed in the literature as a 
method to compress a linear transformation, the building
blocks involved are similar to those used in the ACDC 
structured efficient linear transformation and so we consider it here.

\subsection{Compression Ratio Scaled Weight Decay}
\label{methods:crs}

One way to motivate L2 regularisation in neural networks is to say that it is
equivalent to MAP inference with a Gaussian prior on the
weights~\citep[p.225]{murphy2012machine}. In the context of changing layer structure, we would wish to preserve the total variance 
of the weight matrix prior under the layer replacement.  As the number of parameters tends towards the number in the full weight matrix, we will then tend toward the original weight decay factor. Under the simplifying assumption that the weights, $\{w_n\}_{n=1}^N$ are 
Gaussian distributed with variance $\frac{1}{\sqrt{d}}$, where $d$ 
is the weight decay factor, then total variance is:
\begin{align}
\sum_{n=1}^N E\left[w_n^2 \right] = \sum_{n=1}^N \frac{1}{d} E \left[z^2
\right]\text{,}
\end{align}
where $z$ is Gaussian distributed with variance 1. Then we have:
\begin{align}
    \sum_{n=1}^N Var (w_n) = \frac{N}{d} \text{.}
\end{align}

Hence, to maintain total variance for $M$ parameters in the compressed layer,
we must use the scaled weight decay term: 

\begin{align}
d_c = \frac{M d}{N} \text{.}
\end{align}

In practice this means multiplying the weight decay factor for compressed weight
matrices by the compression ratio $M/N$. In Figure~\ref{fig:crs:wrn}, it is
illustrated that this stabilises training and improves performance. It
has the desirable property of providing a smooth interpolation to an
uncompressed matrix -- where the weight decay would be the
default.  We refer to this approach as \emph{compression ratio scaled} (CRS) weight decay.

\section{Experiments}
\label{experiments}

We train networks where each pointwise ($1\times 1$) convolution in a base network is substituted for a particular linear transform from Section~\ref{methods:linear}. To recaptiulate, these transforms are: ACDC, Tensor-Train, Tucker decomposition, Rank-factorised (RF) decomposition, HashedNet, and Linearised ShuffleNet. Each of these substitutions has a tuning parameter that can be altered to determine the number of parameters utilised, allowing us to compare networks for a range of parameter budgets.
We perform experiments on CIFAR-10~\citep{krizhevsky2009learning} with
(i) Wide-ResNets (WRN)~\citep{zagoruyko2016wide}, specifically WRN-28-10, and (ii) the network discovered by differentiable architecture search (DARTS)~\citep{liu2019darts} as base networks. We also experiment on ImageNet~\citep{russakovsky2015imagenet} with WRN-50-2 as a base network. We chose parameter budgets over which networks with each substitution under consideration
would have support (see Appendix~\ref{appendix:budgets}). When training:

\begin{enumerate}

    \item We perform \emph{attention transfer} (AT)~\citep{zagoruyko2017paying} on each substitute network with a trained version of the original base network as a teacher.
    \item We utilise \emph{CRS weight decay}, as defined in Section~\ref{methods:crs}.
\end{enumerate}

We examine the importance of these choices in Section~\ref{sec:ablation}. Our experimental set up is described in detail in Appendix~\ref{appendix:setup}.

\subsection{WRN-28-10 on CIFAR-10}
\label{experiments:params:wrn}

The base WRN-28-10 achieves a top-1 validation error of 3.2\% and has 36.5M parameters. We produce substitute networks at three approximate parameter budgets: 2.4M, 1.2M, and 0.6M. Note that these correspond to very high compression rates.

\begin{figure}[!h]
   \centering\includegraphics[width=\linewidth]{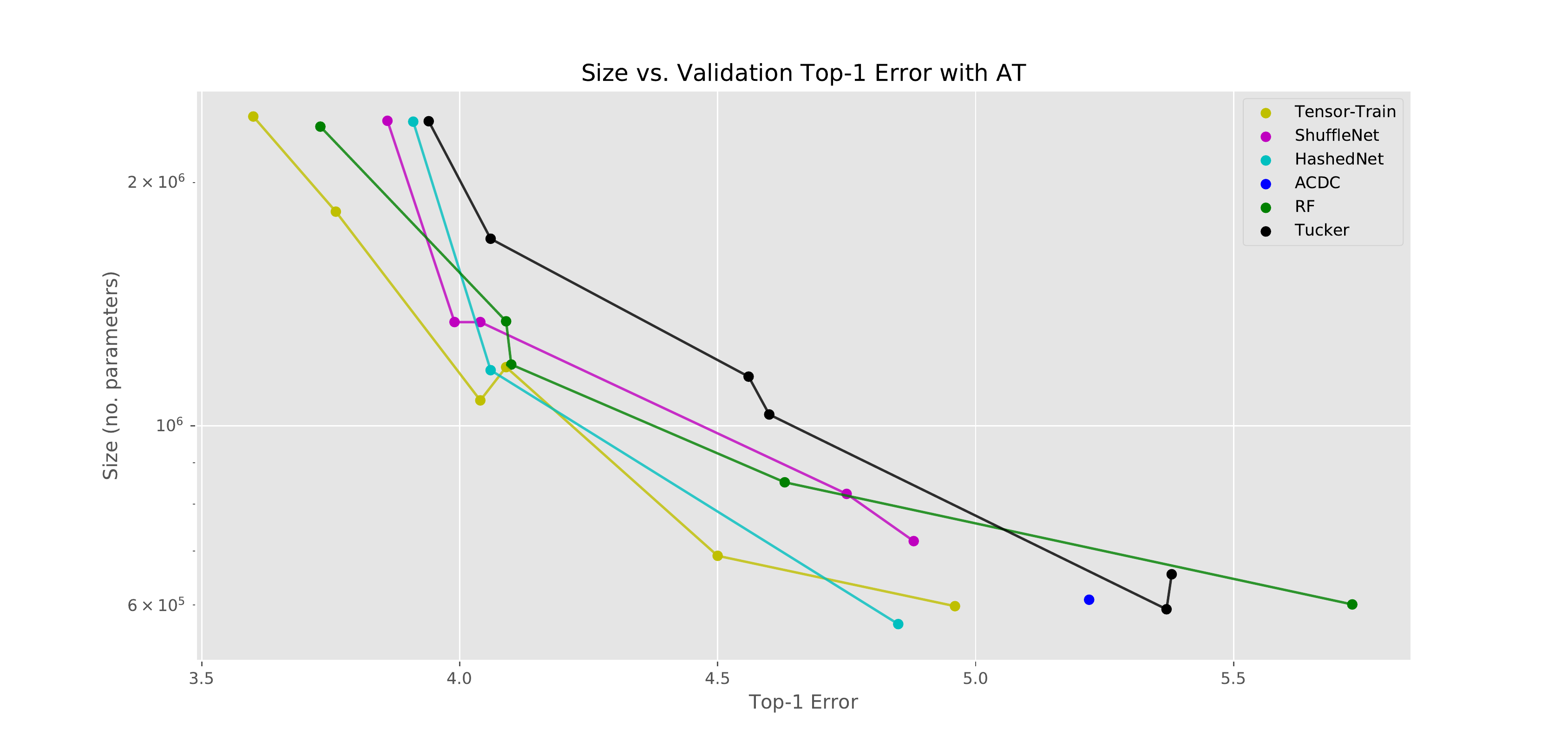}
        \captionof{figure}{The relationship between top-1 error on the validation set
      and the number of parameters is plotted for experiments involving
      WRN-28-10 on CIFAR-10, for experiments using AT.
      Each substitute linear transform tested is
      indicated in the legend. 
      On this problem, both Tensor-Train and HashedNet substitutions are
      able to achieve the highest rates of compression while maintaining performance. At lower compression
      settings, all methods compared achieve comparable top-1 error.}
      \label{fig:wrn:params}
        
\end{figure}

\begin{figure}[!h]
   \centering\includegraphics[width=\linewidth]{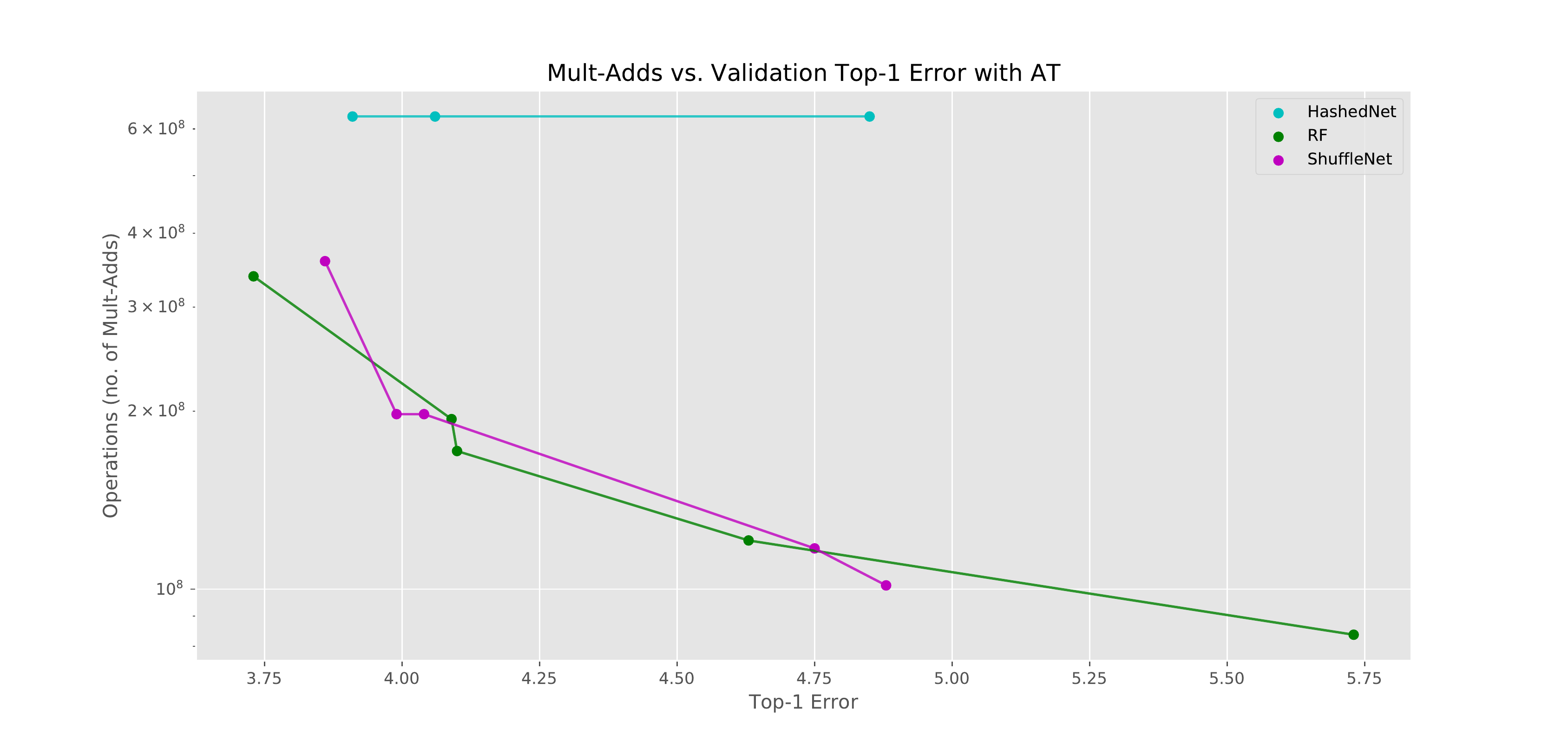}
   \caption{The relationship between top-1 error on the validation set and the number of mult-adds consumed by each network is plotted for experiments involving WRN-28-10 on CIFAR-10. We were unable to calculate robust estimates for the mult-adds used by TT or Tucker substitutions, as they would depend on the choice of rounding and efficient matrix-vector multiplication algorithms used~\citep{oseledets2011tensor}. The only stable ACDC experiment cost $1.8 \times 10^9$ mult-adds, far more than competing methods. Upon investigation, this experiment used a substitution of 12 ACDC layers (as used in the original paper) and this only uses fewer mult-adds than the original linear transform for layers of more than 625 units.}

   \label{fig:wrn:multadds}
\end{figure}

The relationship between the
number of parameters used by our substitute WRN-28-10 networks and the top-1 error---through AT with the base network---is illustrated in Figure~\ref{fig:wrn:params}. We report mult-adds where possible in Figure~\ref{fig:wrn:multadds}.

One might expect HashedNet or Tensor-Train
to work best as compression methods, as they do not necessarily reduce the
number of mult-adds used by the network.  HashedNet, in particular, substitutes
a weight matrix of precisely the same size at test time, and applying that
weight matrix uses the same number of mult-adds used by the original network.
While all of the considered methods produce a network that contains less than
10\% the parameters used by the base network while losing only 1\% error,
HashedNet and Tensor-Train substitutions can maintain this error using \emph{less
than 3\% of the parameters}.
Unfortunately, ACDC was only able to place a single point at the lowest compression
ratio. It performs comparably well, but becomes unstable with larger numbers of
ACDC layers.R

\subsection{DARTS net on CIFAR-10}

\label{experiments:params:darts}

DARTS net is a state-of-the-art image classification network~\citep{liu2019darts} achieving 2.83\% while using only 3.8M parameters. We use this as a base network, and substitute linear transformations for parameter budgets of 1.42M, 0.83M, and 0.49M. As before, these substitute networks are trained using AT with the base network and CRS weight decay.

\begin{figure}[!h]
   \centering
\includegraphics[width=\linewidth]{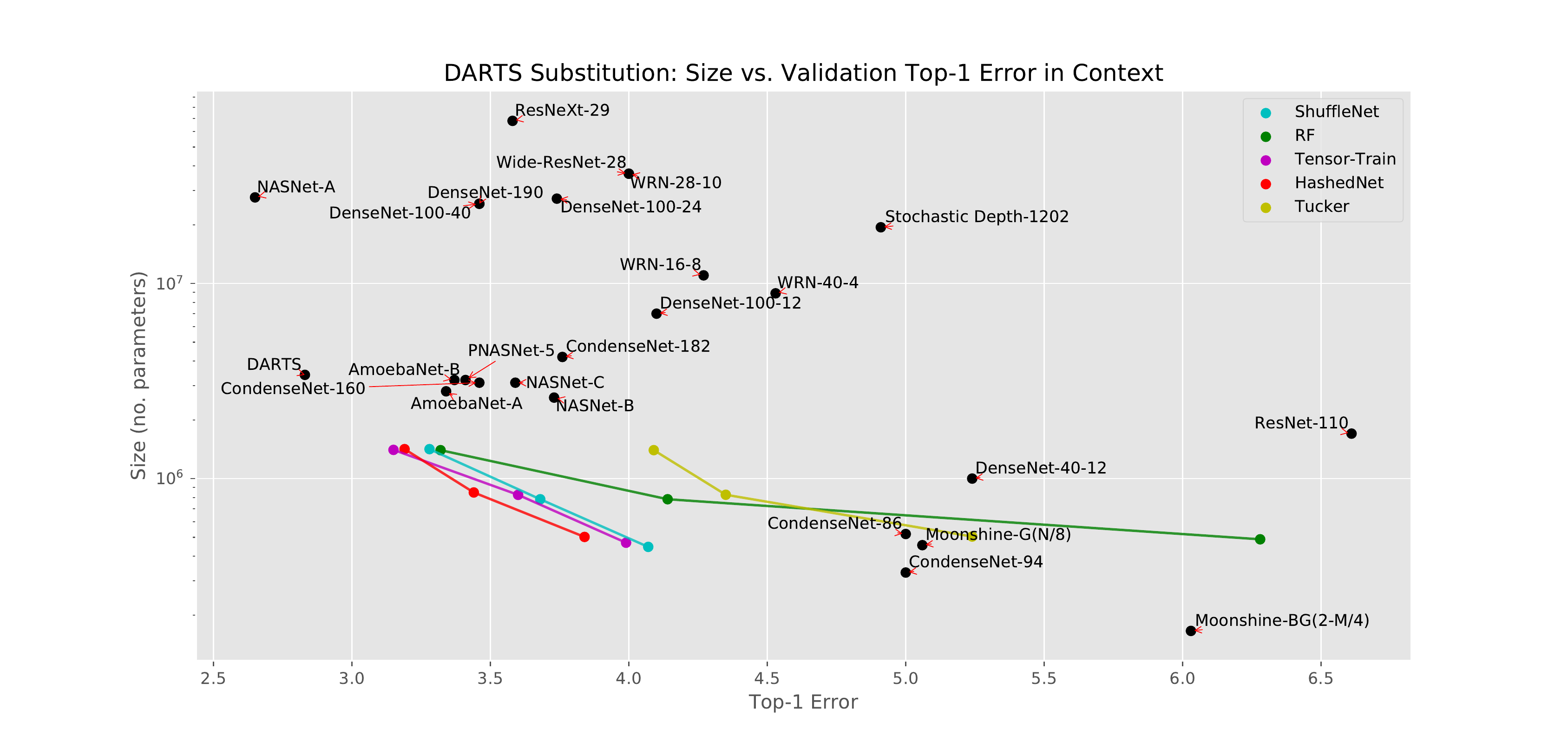}
  \caption{Top-1 validation errors on CIFAR-10 for DARTS networks with substitute linear transforms and their parameter totals. Each substitute linear transform tested is
  indicated in the legend. ACDC is omitted as it failed to converge below 8\% in any case. We compare against recent networks presented in the
  literature, including: DenseNet~\citep{huang2017densely},
  Moonshine~\citep{crowley2018moonshine}, Wide ResNet~\citep{zagoruyko2016wide},
  ResNeXt~\citep{xie2017aggregated}, Stochastic Depth~\citep{huang2016deep},
  GoogleNet~\citep{szegedy2015going}, CondenseNet~\citep{huang2018condensenet},
  NASNet~\citep{zoph2018learning}, ResNet~\citep{he2016deep},
  PNASNet~\citep{liu2018progressive}, AmoebaNet~\citep{real2018regularized} and
  DARTS~\citep{liu2019darts}. \emph{Using these substitutions  we are able to explore a new region in the Pareto Frontier.}}
  \label{fig:darts}
\end{figure}

The validation errors of these substitute networks against their parameter total is shown in Figure~\ref{fig:darts}. Remarkably, we can achieve compression to 20\% of the original number of
parameters for HashedNet, ShuffleNet and Tensor-Train substitutions while still being {\it within 1\% the original
top-1 error}.

\begin{figure}[!h]
   \centering\includegraphics[width=\linewidth]{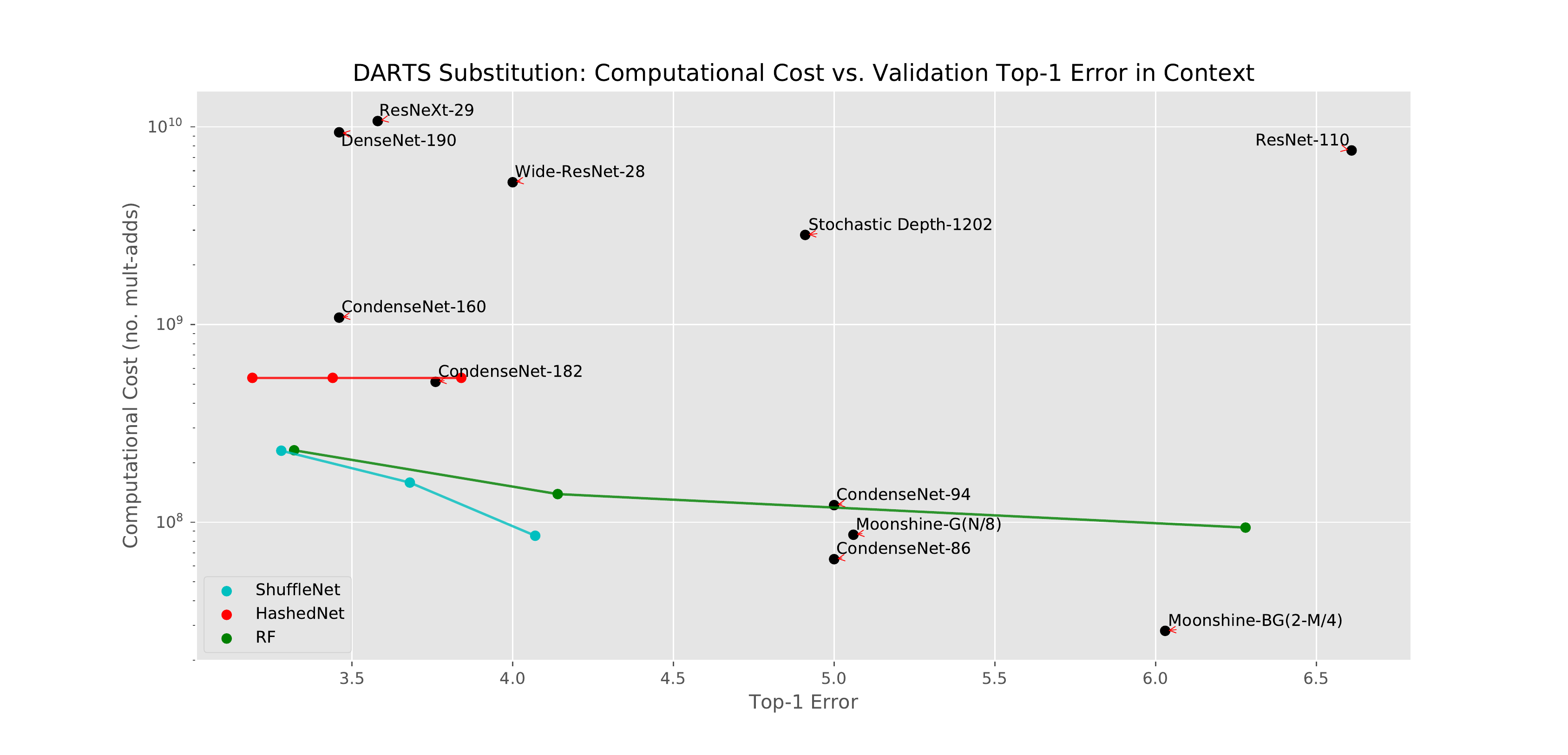}
   \caption{Top-1 validation errors on CIFAR-10 for DARTS networks with substitute linear transforms and their associated mult-add cost. Not all substitute transforms tested could be included here, as noted in Figure~\ref{fig:wrn:multadds}, not all could be easily estimated. We compare against recent networks in the literature, including: ResNext~\citep{xie2017aggregated}, DenseNet~\citep{huang2017densely}, Wide ResNet~\citep{zagoruyko2016wide}, CondenseNet~\citep{huang2018condensenet}, Stochastic Depth~\citep{huang2016deep}, ResNet~\citep{he2016deep} and Moonshine~\citep{crowley2018moonshine}. On this Figure, \emph{ShuffleNet substitutions appear to be Pareto optimal}, but we were not able to compare against a large number of papers that do not report mult-add cost on CIFAR-10.}
   \label{fig:darts:multadds}
\end{figure}

We can see that our networks  conveniently explores an {\it empty region of the Pareto frontier} in the context set
by the literature. The top-1 error achieved through a HashedNet substitution is equal
to or lower than all published networks compared against, save for DARTS and
NASNet-A, while using several times fewer parameters. 

When we compute the mult-adds used by these networks (Figure~\ref{fig:darts:multadds}) we observe similar trends. Notably, our ShuffleNet substitution performs extremely similarly to the original network while using around 5 times fewer operations. In terms of mult-adds, these networks again extend the Pareto boundary defined by all methods considered.

\subsection{WRN-50-2 on ImageNet}
\label{experiments:params:imagenet}

Based on their performance
in the two CIFAR-10 experiments, we chose HashedNet, Tensor-Train and
ShuffleNet to compare on ImageNet with WRN-50-2 as a base network. We also included RF substitution as a baseline.

The results of the experiments are shown in Table~\ref{table:imagenet}.
ImageNet is a more difficult problem than CIFAR-10, and we see that performance
rapidly degrades as we reduce the number of parameters, although this appears
to be the same trend observed with published networks in the literature. The compression rates achieved with our agnostic substitutions compare
favourably to other state-of-the-art image classification networks in the
field. This demonstrates the generalisation of this method to even the largest
deep convolutional neural networks for image classification.

\begin{table}[!h]
    \caption{Top-1 validation errors on ImageNet for WRN-50-2 with our proposed substitutions: ShuffleNet, Tensor-Train and RF. Each method is tested
        at two approximate parameter budgets.
        Compression is given as a percentage of the original
        model size. Methods from the literature are provided for comparison:
    WRN-50-2~\citep{zagoruyko2016wide}, ShuffleNet~\citep{zhang2018shuffle},
    DenseNet~\citep{huang2017densely}, 
    MobileNet~\citep{howard2017mobilenets},
    DecomposeMe~\citep{alvarez2017compression}, 
    ACDC~\citep{moczulski2016acdc} and TT~\citep{novikov2015tensorizing}.}
    \small
    \centering
    \begin{tabular}{p{0.13\textwidth}p{0.12\textwidth}rrp{0.1\textwidth}p{0.07\textwidth}p{0.07\textwidth}}
        \toprule
              &              &            &           &            & \multicolumn{2}{c}{Compression (\%)} \\
        Model & Substitution & Parameters & Mult-Adds & Top-1 (\%) & Parameters & Mult-Adds \\
        \midrule
        WRN-50-2 & ShuffleNet            & 6.04M       & 0.91G      & 29.7          & 8.77            & 8.00           \\
        WRN-50-2 & ShuffleNet            & 17.72M      & 3.22G      & 26.9            & 25.72           & 28.22          \\
        WRN-50-2 & RF & 4.35M       & 0.53G      & 39.8           & 6.3            & 4.62           \\
        WRN-50-2 & RF & 17.55M      & 2.83G      & 25.4            & 25.5           & 24.77          \\
        WRN-50-2 & HashedNet             & 4.35M       & 4.86G      & 33.5            & 6.32            & 42.59          \\
        WRN-50-2 & HashedNet             & 17.61M      & 4.86G      & 24.5            & 25.56           & 42.59          \\
        WRN-50-2 & Tensor-Train          & 4.34M       & 4.86G      & 33.2            & 6.30            & 42.59          \\
        WRN-50-2 & Tensor-Train          & 17.58M      & 4.86G      & 24.9            & 25.52           & 42.59          \\
    	\hline
        \multicolumn{2}{p{0.3\textwidth}}{WRN-50-2}       & 68.9M & 11G   & 21.9 & & \\
        \multicolumn{2}{p{0.3\textwidth}}{ShuffleNet}     & 1.87M & 0.14G & 32.4 & & \\
        \multicolumn{2}{p{0.3\textwidth}}{ShuffleNet 2x}  & 7.51M & 0.53G & 24.7 & & \\
        \multicolumn{2}{p{0.3\textwidth}}{DenseNet-121}   & 9M    & 6G    & 25.0 & & \\
        \multicolumn{2}{p{0.3\textwidth}}{MobileNet}  & 4.2M  & 0.57G & 29.4 & & \\
        \hline        
        \multicolumn{2}{p{0.3\textwidth}}{$\text{Dec}_{8}^{512}$} & & 0.45G & 33.2 & 46.5 & 53.8 \\
        \multicolumn{2}{p{0.3\textwidth}}{CaffeNet(ACDC)} & 9.7M & & 43.3 & 16.7 & \\
        \multicolumn{2}{p{0.3\textwidth}}{VGG-16(TT)} & 18.65M & & 32.2 & 13.5 & \\
        \multicolumn{2}{p{0.3\textwidth}}{VGG-19(TT)} & 24.0M & & 31.6 & 16.7 & \\
        \bottomrule
    \end{tabular}
    \label{table:imagenet}
\end{table}

\section{Ablation Studies}
\label{sec:ablation}

\subsection{Training With Distillation}
\label{experiments:distil}

\begin{figure}
   \centering\includegraphics[width=\linewidth]{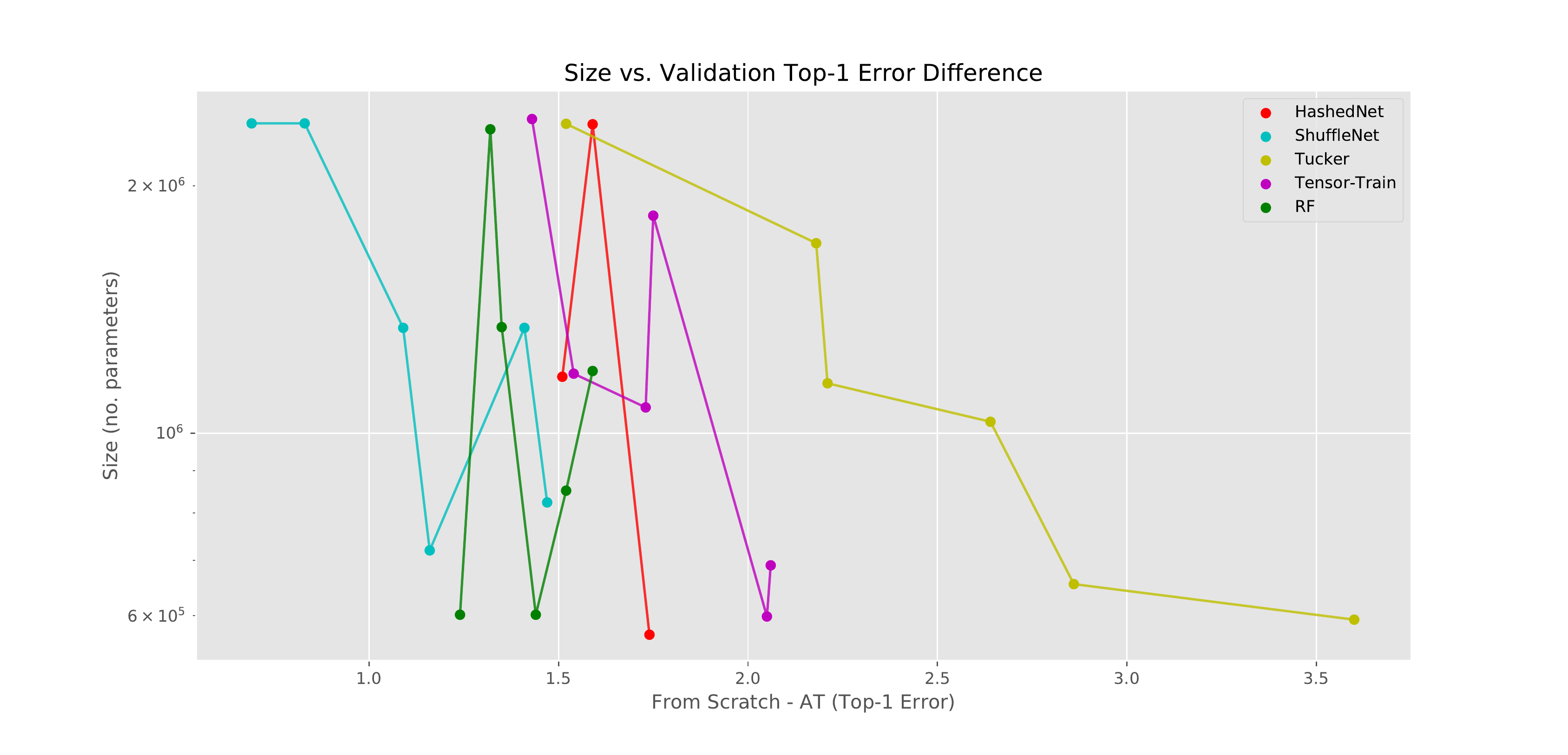}
        \captionof{figure}{The difference in top-1 validation error with and without AT
        distillation is plotted for each substitute linear transform applied to
        WRN-28-10 on the CIFAR-10 classification problem. Over all the linear
        transform substitutions, AT lowers the top-1 validation error by a few
        percent.  It has the greatest benefit in conjunction with the Tucker
        substitution, which has a higher overall top-1 validation error, as can be
        seen in Figure~\ref{fig:wrn:params}.}
        \label{fig:wrn:at:ablation}
\end{figure}

Distillation via attention transfer~\citep{zagoruyko2017paying}---or AT---is used for most of our experiments because we saw it universally improved top-1 validation
error in the experiments with WRN-28-10 on CIFAR-10.
Figure~\ref{fig:wrn:at:ablation} illustrates the difference in top-1 error
validation error in the WRN-28-10 experiments.  Almost all methods benefit by
more than 1\%, which can be critical for a competitive top-1 error score on
CIFAR-10. The ACDC
substitution was omitted, as it was not stable with AT (but was also not competitive regardless).

Unlike methods, such as a hyperparameter search, that allow us to spend more
computation time to achieve a better top-1 error, AT does not require tuning. We
did not tune any hyperparameters in any of the experiments presented here. For
this reason, we can view AT as a way to compensate for an inadequate training
routine.

\subsection{CRS Weight Decay Ablation}
\label{ablation:crs}

\begin{figure}[!h]
    \centering
    \includegraphics[width=0.8\textwidth]{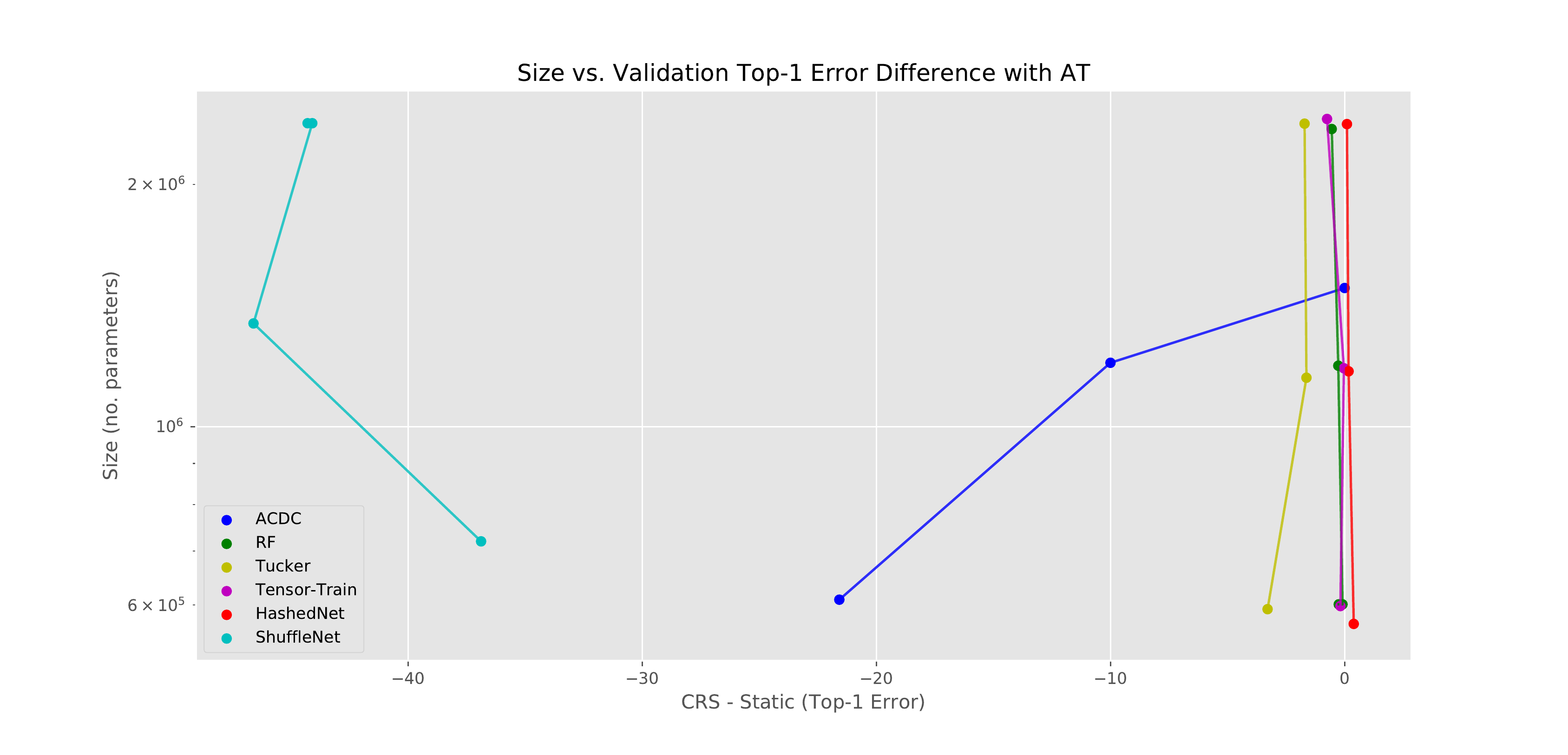}
    \caption{The difference in top-1 error on the validation set of CIFAR-10, when
    training with and without CRS weight decay, over all the substitution
    methods considered for WRN-28-10. For all methods apart from HashedNet, this form
    of weight decay scaling is beneficial; it results
    in a lower top-1 validation error.}
    \label{fig:crs:wrn}
\end{figure}

To justify CRS weight decay, we ran an ablation experiment, repeating the
experiments on CIFAR-10 with WRN-28-10, but disabling CRS weight decay.  In
Figure~\ref{fig:crs:wrn} these results are illustrated. For almost all methods
we see that there is a clear benefit.  ShuffleNet simply fails to converge
without it. However, for HashedNet we see that it is slightly detrimental.

\begin{figure}[!h]

    \centering
    \includegraphics[width=\textwidth]{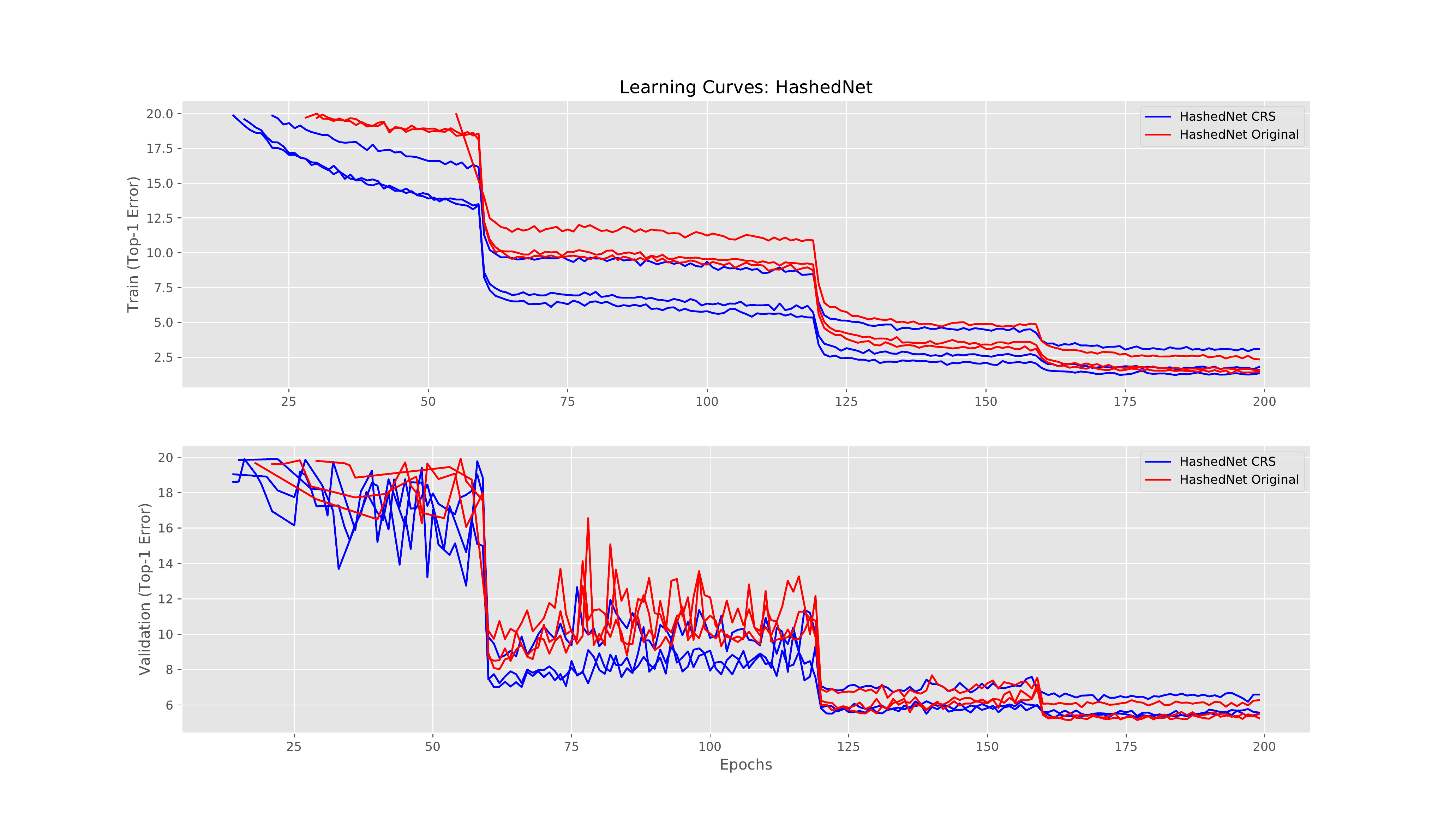}

    \caption{Learning curves for HashedNet substitution experiments, with
    and without CRS weight decay. When CRS weight decay is enabled
    the top-1 error is lower, on train and test, at every epoch until
    the final part of the learning rate schedule.}
    \label{fig:crs:hashedlc}

\end{figure}

To investigate why this happens, Figure~\ref{fig:crs:hashedlc}
illustrates the learning curves---top-1 error plots against 
current training epoch---of these HashedNet substitute 
networks. The CRS weight decay stabilises training as we would
hope, and the top-1 validation error is lower with
it enabled \emph{until the final stage} of the learning rate schedule.
At this stage we can see the training top-1 error decreases
faster when CRS weight decay is enabled. This overfitting is
enough to cause a slight increase in top-1 error.

\section{Conclusion}
\label{conclusion}

Many alternative efficient dense layers have been proposed in prior work.  In
this paper we have reviewed a selection of these layers to be substituted into state-of-the-art image classification networks. This was motivated by the
potential efficiency benefits, at training and test time, of using such layers
in real applications. Training using such layers as substitutes in existing image classification
architectures can be trivially stabilised by using CRS weight decay; a rote
method to scale the weight decay prescribed for the original architecture. As
this does not require hand tuning, it allows many alternatives to expensive
convolutional layers to be used. We found that these substitutions were capable of performing efficiently on
contemporary image classification benchmarks.  We have explored a new region of the Pareto frontier for both parameter and computational cost on CIFAR-10. On ImageNet, we find it is possible to match
the compression performance of many published methods, all while using a
network in which the weights are a linear reparameterisation of the original.

\paragraph{Acknowledgements}

\noindent This study was supported in part by an EPSRC scholarship granted to GG from the
Neuroinformatics and Computational Neuroscience Doctoral Training Centre at the University of Edinburgh, a Huawei DDMPLab Innovation Research Grant, as well as funding from the European Union's Horizon 2020 research and innovation programme under grant agreement No.~732204 (Bonseyes). This work is supported by the Swiss State Secretariat for Education, Research
and Innovation (SERI) under contract number 16.0159. The opinions expressed and
arguments employed herein do not necessarily reflect the official views of
these funding bodies. The authors are grateful to Rafael Ballester for correspondence on tensor-train practicalities, and to Joseph Mellor for productive discussions on the CRS weight decay scheme and HashedNet.

\FloatBarrier
\bibliography{main}
\bibliographystyle{icml2019}

\appendix
\clearpage

\section{HashedNet Disconnected Weights}
\label{appendix:hashednet}

The indices produced by the hash function are approximately uniform over the
set of real weights. This produces a weight matrix in which weights are
randomly tied, with each unique weight occurring on average the same number of
times. \citet{chen2015compressing} demonstrate that the cost of accessing these
weights is negligible at test time. In our experiments, we do not use a hash
function, instead sampling the indices once when the layer is initialised and
storing them. 

The number of parameters to be optimised here is the number of~\emph{real}
weights $\wB$, which can be set to be 1 or greater, up to the number of elements
in the virtual weight matrix. However, as the number of real weights is
increased the probability we may store a weight that is never used in the virtual
weight matrix increases. If $N_r$ is the number of real weights and $N_v$ is the
number of virtual weights, then the expected number of weights that will be
excluded will be $N_r (1 - 1/N_r)^{N_v}$. Defining $N_r$ in terms of $N_v$ using
a compression ratio $c = \frac{N_r}{N_v}$, we can investigate what happens to
the ratio excluded, $e$, as $c$ changes:
\begin{align}
    e & = \left( 1 - \frac{1}{c N_v} \right)^{N_v} \\
      & = \exp \left( N_v \log \left( 1 - \frac{1}{c N_v} \right) \right).
\end{align}

Taking the Taylor expansion of $\log \left( 1 - \frac{1}{c N_v} \right) $
and retaining the first two terms:
\begin{align}
    e & = \exp \left( N_v \log \left( 1 - \frac{1}{c N_v} \right) \right) \\
      & = \exp \left( N_v \left( -\frac{1}{c N_v} + O(\frac{1}{(c N_v)^2}) \right) \right) \\
      & \approx \lim_{N_v \to \inf} \exp \left( N_v \left( -\frac{1}{c N_v} + O(\frac{1}{(c N_v)^2}) \right) \right) \\
      & = \exp \left( -\frac{N_v}{c N_v} \right) \\
      & = \exp \left( -\frac{1}{c} \right).
\end{align}

\begin{figure}
    \centerline{
        \includegraphics[width=0.7\textwidth]{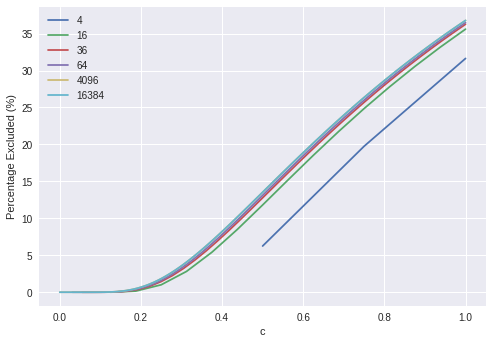}
	}
    \caption{The effect on percentage of weights excluded depending on 
    compression ration $c$, tested for different values of $N_v$, the
    number of elements in the virtual weight matrix, indicated in the
    legend. At the compression levels we are interested in---20\% of
    the original number of weights---we can see that the number of weights
    excluded is low.}
    \label{fig:hashed-excluded}
\end{figure}

As shown in Figure~\ref{fig:hashed-excluded}, this limit argument holds true for the
values of $N_v$ we are interested in, and the proportion of weights excluded as
the compression ratio grows can be significant. In our experiments we do not address
these wasted parameters, despite performing experiments with compression ratios in
regions where 10-20\% of our parameters are being excluded. It would also be possible
to identify these parameters and choose not to store them, but we do not investigate
this. The reason being that we find the HashedNet substitution
effective at high compression levels, such as below $c=0.1$, and in this region a
negligible number of weights will be excluded. 

\section{Chosen Parameter Budgets}
\label{appendix:budgets}

\begin{figure}[!h]
  \centering\includegraphics[width=0.6\linewidth]{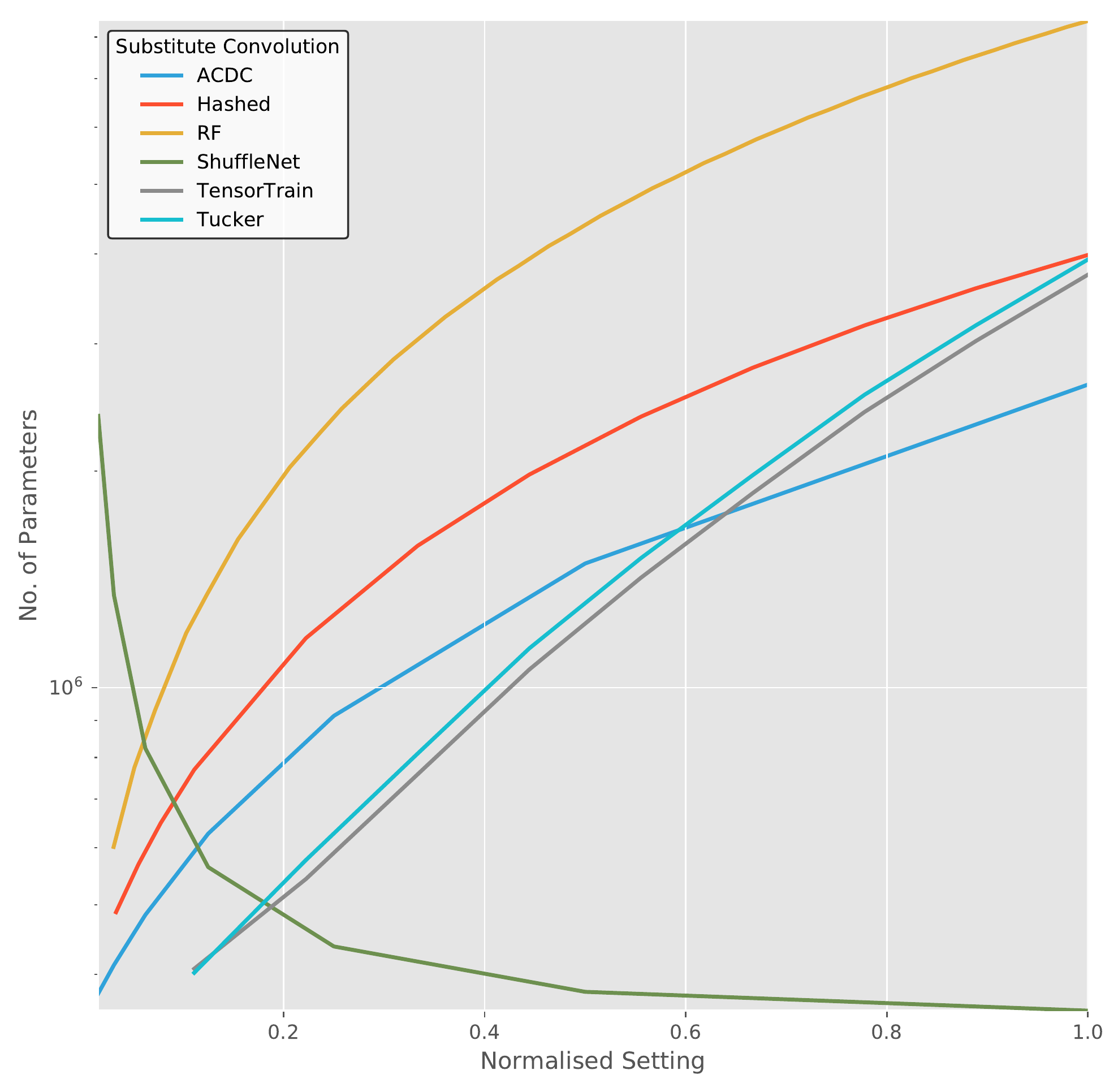}
  \caption{The parameter cost of a WRN-28-10 after substitution by the methods
   listed in the legend, varying the tunable parameter of each over a normalised
   range. We design experiments over a parameter count range such that all methods
   illustrated will have support, which here is limited by the maximum size
   of the Linear ShuffleNet and the minimum size of the RF substitution.}
  \label{fig:wrn:paramcount}
\end{figure}

After normalising the tuning of all layers between 0 and 1, we can
plot number of parameters used by each substitution as shown in 
Figure~\ref{fig:wrn:paramcount}. The upper limit and lower limits 
were chosen where all methods have support. For example, we can
see in Figure~\ref{fig:wrn:paramcount} we can see that the upper
limit is defined by the Linear ShuffleNet, while the lower limit
is defined by RF. We chose the midpoint by linear interpolation
in log parameter count.

\section{Experiment Setup}
\label{appendix:setup}

All experiments were written in Python using PyTorch~\citep{pytorch}.
Tensor-Train and Tucker decompositions were implemented using 
tntorch~\citep{tntorch}; all other methods were implemented 
separately. The code implementing ACDC layers is based on the work of \citet{pytorchdct}, and is publicly available: \url{https://github.com/gngdb/pytorch-acdc}. Figures were produced using Matplotlib~\citep{matplotlib}
and Holoviews~\citep{stevens2015holoviews}.
Annotations on figures were placed using adjustText~\citep{adjusttext}.

CIFAR-10 is a set of 60,000 colour images of size 32 by 32 pixels, with the task
of classifying each image according to 10 classes~\citep{krizhevsky2009learning}.  ImageNet is a
dataset of over a million colour images of size 224 by 224, with the task of
classifying each into 1000 classes~\citep{russakovsky2015imagenet}. The results on CIFAR-10
typically inform experiments planned on ImageNet, which is used as verification
that the method scales to large problems.

For CIFAR-10 experiments we focused on two
architectures: Wide ResNets~\citep{zagoruyko2016wide} (WRN) and the network found in~\cite{liu2019darts} (DARTS).  Wide
ResNets were chosen to demonstrate results on a common ResNet structure. Results
on this type of network should be reflected in many similar networks in the
literature.  Wide ResNets are defined by their \emph{depth} and \emph{width}
factors. We choose to focus on the WRN-28-10, $\text{depth}=28$ and
$\text{width}=10$ in~\cite{zagoruyko2016wide}.

Wide ResNet architectures were used to demonstrate the
results of attention transfer~\citep{zagoruyko2017paying}, 
and we run these networks
using that training protocol. When using attention transfer $\beta$ was set to 1000. 

DARTS was selected in order to demonstrate results on a
state-of-the-art image classification architecture.
We replicated precisely the training
hyperparameters and schedule used in the original paper.

\paragraph{Wide ResNet} Each network was trained for 200
epochs with a learning rate starting at 0.1 and
scaled by 0.2 on epochs 60, 120 and 160. Momentum was set
to 0.9 and the minibatch size was 128.
Weight decay was set to
$5 \times 10^{-4}$ and scaled in all experiments according
to the method described in Section~\ref{methods:crs},
apart from the ablation experiment described in 
Section~\ref{ablation:crs}. Data was augmented with random crops,
left-right flips and Cutout~\citep{devries2017cutout}.

\paragraph{DARTS} Each network was trained for 600 epochs
using a cosine annealed learning rate schedule starting at 0.025.
Momentum was set to 0.9 and the minibatch size was 96. Weight
decay was set to $3 \times 10^{-4}$ and scaled in all experiments according
to the method described in Section~\ref{methods:crs}.
The auxiliary classification head was used in training, but
not counted at test time, and the drop-path method from
the paper followed the same schedule of a linear increase
in drop probability from 0 to 0.2 over the learning schedule.
Data was again augmented with random crops,
left-right flips and Cutout~\citep{devries2017cutout}.

In ImageNet experiments we focused on a large network with
competitive results, in order to demonstrate 
the potential for compression. As with previous experiments
we chose a Wide ResNet~\citep{zagoruyko2016wide}, 
so that results could be interpreted
as transferable to other ResNet-like architectures in the literature.
All ImageNet experiments use the WRN-50-2, which is precisely
a ResNet-50~\citep{he2016deep} with twice as many channels on inner bottlenecks.
The published performance of 21.9\% top-1 error is competitive with
the best published results on ImageNet. We used the
publicly available model zoo trained weights and found it could
only achieve 22.5\%, but chose to use this network regardless.

Each network was trained for 90 epochs with a learning rate of 0.1
scaled by 0.1 at epochs 30 and 60. Momentum was set to 0.9 and 
the minibatch size was 256. Weight decay was $1 \times 10^{-4}$
and scaled according to the method described in 
Section~\ref{methods:crs}. Data was augmented with random
crops and left-right flips.

\end{document}